\documentclass{article}
\usepackage{spconf,amsmath,graphicx, amsmath, amssymb, multirow}
\usepackage{hyperref}       

\title{Binarized Weight Error Networks With a Transition Regularization Term}
\name{Savas Ozkan \thanks{Email: ozkan.savas@metu.edu.tr}, Gozde Bozdagi Akar}
\address{Department of Electrical and Electronics Engineering, Middle East Technical University, \\ 06800, Ankara, Turkey}
\begin{document}
%
\maketitle
\begin{abstract}
\vspace{-0.1cm}

This paper proposes a novel binarized weight network (BT) for a resource-efficient neural structure. The proposed model estimates a binary representation of weights by taking into account the approximation error with an additional term. This model increases representation capacity and stability, particularly for shallow networks, while the computation load is theoretically reduced. In addition, a novel regularization term is introduced that is suitable for all threshold-based binary precision networks. This term penalizes the trainable parameters that are far from the thresholds at which binary transitions occur. This step promotes a swift modification for binary-precision responses at train time. The experimental results are carried out for two sets of tasks: visual classification and visual inverse problems. Benchmarks for Cifar10, SVHN, Fashion, ImageNet2012, Set5, Set14, Urban and BSD100 datasets show that our method  outperforms all counterparts with binary precision.

\end{abstract}
\vspace{-0.3cm}

\section{Introduction}
\label{sec:intro}
\vspace{-0.3cm}

Over the past decade, neural networks have shown remarkable performance in various fields~\cite{he2016deep, simonyan2014very}. In particular, this structure transforms high-dimensional data into low-dimensional latent space by taking advantage of non-linear transformations and vast parameters. However, in addition to its popularity, this competence also multiplies the computation load. Increasing computational and storage demands, therefore, inevitably reduce its applicability on portable devices.

In recent years, several critical studies aim to reduce the need for these limitations with a slight decrease in performance than full precision counterparts. In general, compressing full precision parameters into binarized ones plays a prominent role because of the simplicity and efficiency
~\cite{soudry2014expectation, courbariaux2015binaryconnect, courbariaux2016binarized, rastegari2016xnor, hubara2017quantized, zhou2016dorefa}. Mathematically, this type of methods minimizes the approximation error between full and binary weights as follows: 
\begin{eqnarray}
\label{eqn:bc1}
J(\alpha, \mathbf{W}^b) = \underset{\alpha, \mathbf{W}^b}{\mathrm{argmin}}{|| \mathbf{W} - \alpha \mathbf{W}^b ||}
\end{eqnarray}

\noindent Here, $\mathbf{W}$ and $\mathbf{W}^b$ indicate full and binary precision trainable parameters, respectively. Moreover, $\alpha$ denotes a scale factor where its optimal solution is equal to mean of the absolute value of full precision weights $\mathbb{E}[|\mathbf{W}|]$. In the end, this approximation leads to $\mathbf{W}\approx\alpha \mathbf{W}^b$ with a significant quantization error. Observe that this quantization error is even worse for shallow networks since the allowable range for parameters is large. This issue eventually weakens the stability of solutions. Additionally, error propagation for binary precision parameters can be severe due to limited operations in back-propagation steps (i.e., multiplication with binarized weights instead of double precision). This limitation causes additional problems for converging optimal solutions.

In this work, the quantization error in Eq.~\ref{eqn:bc1} is reduced by an iterative quantization step. In particular, full weights are iteratively quantized in several steps, and a binary representation is computed with a negligible increase in the computational load. As noticed, this schema intuitively shares an assumption similar to hierarchical clustering, which tries to create a hierarchy of clusters. Similarly, this step helps improve the stability of binarized weights because of the better partitions. Indeed, our bitwise operations induce a fast inference similar to other quantized neural network by consuming less power and memory (appx. $16\times$) on the devices. However, performance can slightly reduce due to the bit-length compared to full-precision counterparts.

In addition, a regularization term is introduced that is feasible for all threshold-based binarized networks. This term penalizes parameters that are far from the thresholds at which binary transitions occur. Hence the parameters are concentrated close to the binary transitions. This term favors changing weights quickly in binary precision to observe their effects to the performance and avoid overfitting for parameters.

The rest of the paper is formed as follows. First, the related work for binarized weight networks is summarized. Later, the concept of binarized weight networks and our contributions are described. Lastly, we explain the test results and final remarks.

\vspace{-0.3cm}
\section{Related Work}
\label{sec:related}
\vspace{-0.3cm}

Although there is a large body of studies related to the compression of computation and memory requirements for deep neural networks~\cite{iandola2016squeezenet}, binary-precision networks play a crucial role due to their efficiency and simplicity~\cite{soudry2014expectation, courbariaux2015binaryconnect}. 

In baseline methods, all trainable parameters are binarized with a quantizer. Later, the backpropagation step is also realized on these binary-precision weights. \cite{rastegari2016xnor} shows that the performance for binarized-parameters can be further improved by adding a single scale factor. Moreover, activation and parameters are also converted to binary-precision to achieve a binarized inference~\cite{rastegari2016xnor, hubara2017quantized, zhou2016dorefa}. 

Note that these methods limit weights to \{-1, 1\} (with some scale factor) and decrease the sparsity property of neural networks. \cite{li2016ternary, zhu2016trained} explain that the use of zero weights \{-1,0,1\} contributes to performance by promoting sparsity in their solutions. In this way, robust representations can be computed. \cite{zhu2016trained} also shows that multiple scale factors obtain better results compared to the cases where a single scale factor is utilized. Recently, SBNN prunes some of NN layers in a binarized network and refines the network with a softsign function~\cite{wu2020sbnn}. The most similar work~\cite{yang2019quantization, zhang2018lq} to our method is to quantize continuous parameters to a set of integer numbers with multiple binary quantizers. However, this can lead to overly quantized representations where instability occurs. We recommend~\cite{qin2020binary, roth2020resource} to read for the survey in binary neural networks.

Lastly, the applications of this terminology are proliferated on different tasks~\cite{ma2019efficient, chiang2020deploying}.

\vspace{-0.3cm}
\section{Binarized Weight Error Networks}
\label{sec:proposal}
\vspace{-0.3cm}

\begin{figure}
\centering
\includegraphics[scale=0.13]{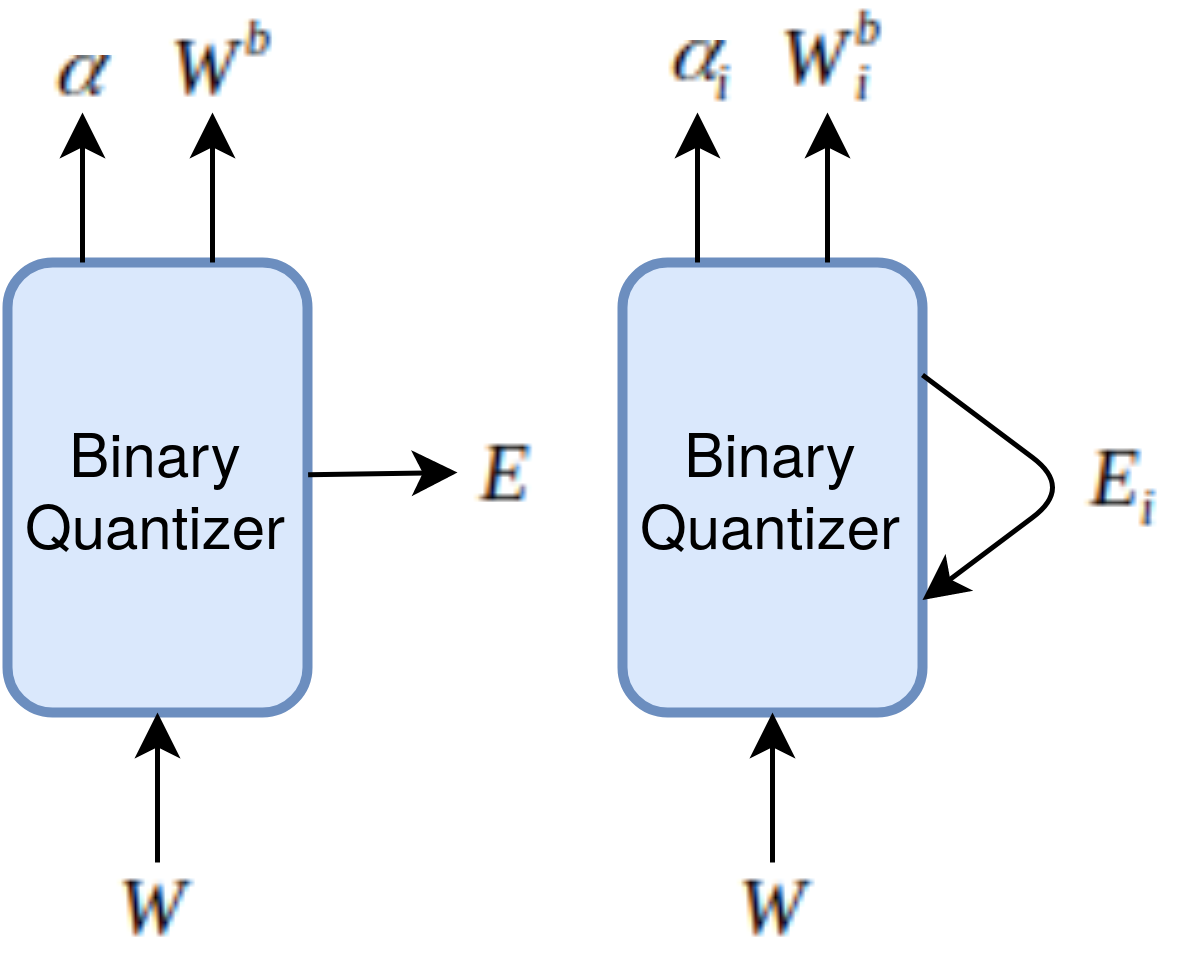}

\vspace{-0.3cm}
\caption{The flow of conventional (first) and our method (second) for the binarized network. Differently, our method quantizes full-precision parameter $W$ by counting approximation error $E$.}
\vspace{-0.4cm}

\label{fig:f0}
\end{figure}

The motivation of binarized neural networks is to minimize the error between full-precision $\mathbf{W}$ and binary-precision $\mathbf{W}^b$ weights as formulated in Eq.~\ref{eqn:bc1}. Generally, two binarized network structures come into prominence in the literature. 

In binary-weight networks, an approximation problem is solved by assigning a binary weight as:
\begin{eqnarray}
\label{eqn:bc2}
\mathbf{W}^b = \begin{cases}
+1, & \text{if } \mathbf{W}\geq 0 \\
-1, & \text{if } \mathbf{W}<0 \end{cases}
\end{eqnarray}

Implicitly, this overall solution  corresponds to the operation $\mathbf{W}^b=\text{sign}(\mathbf{W})$. Note that $\mathbf{W}^b$ takes only $\{-1,1\}$, and the responses of weights are always active since there is no zero operation. This assumption conflicts with the conclusions reached about filter characteristics, as explained in~\cite{krizhevsky2012imagenet} (i.e., Gabor-type filtering and sparsity).

On the contrary, ternary-weight networks compute a representation with a threshold function in Eq.~\ref{eqn:bc2}, since there is no deterministic solution for binary-weights as claimed in~\cite{hwang2014fixed}:
\begin{eqnarray}
\label{eqn:bc2}
\mathbf{W}^b = \begin{cases}
+1, & \text{if } \mathbf{W}>\Delta \\
 0, & \text{if } |\mathbf{W}|<\Delta \\
-1, & \text{if } \mathbf{W}<-\Delta \end{cases}
\end{eqnarray}

\noindent Here, this assumption is exploited to approximate the threshold value $\Delta$ as $0.66\cdot\mathbb{E}[|\mathbf{W}|]$. This model introduces more representation capacity because of the sparsity,  while stability is also increased. 

\vspace{-0.3cm}
\subsection{Binarized Weight Error Networks}
\label{ssec:weighterr}
\vspace{-0.1cm}

As excepted, quantization errors for binary weights are moderately high compared to ternary weights. However, we believe there is room to take additional steps to reduce this error.

In particular, the idea emphasized in our work is based on the fact that an iterative approach can be more suitable for binarization process to increase the capacity of representations than the quantization of full-precision parameters in a single step. For this purpose, the problem definition in Eq.~\ref{eqn:bc1} is rewritten as:
\begin{eqnarray}
\label{eqn:bc3}
J_i(\alpha, \mathbf{W}_i^b) = \underset{\alpha, \mathbf{W}_i^b}{\mathrm{argmin}}{|| \mathbf{E}_i - \alpha_i \mathbf{W}_i^b ||}
\end{eqnarray}

\noindent Here, $\mathbf{E}_i$ indicates the quantization error that occurs in the previous iteration step $i-1$. This is equal to $\mathbf{E}_0=\mathbf{W}$ in the first iteration. When the total iteration is set to 2, the final representation can be approximated as:
\begin{eqnarray}
\label{eqn:bc4}
\mathbf{W}^b  = \alpha_{opt} \cdot (\mathbf{W}_0^b + \mathbf{W}_1^b)
\end{eqnarray}

\noindent Here, $\alpha_{opt}$ indicates the optimum scale value for this conversion that needs to be recalculated. The flows of conventional and our method for binarized networks are illustrated in Fig.~\ref{fig:f0}.

An observation for Eq.~\ref{eqn:bc4} is that the use of two binary-weight models for both $\mathbf{W}_0^b$ and $\mathbf{W}_1^b$ implicitly reproduces a ternary-weight model. This result is critical because it confirms the mathematical proof of our method. Note that this step also shows that binary-representation can be improved further with simple replacements in the formulation.   

Second, we compute the optimum scale value $\alpha_{opt}$ as: 
\begin{eqnarray}
\label{eqn:scale}
\alpha_{opt} = 0.75\cdot \alpha_1 - 0.25\cdot \alpha_2
\end{eqnarray}

\noindent where $\alpha_1$ and $\alpha_2$ are equal to $\mathbb{E}[|\mathbf{E_0}|]$ and $\mathbb{E}[|\mathbf{E_1}|]$, respectively. If binary and ternary models are sequentially applied by using Eq.~\ref{eqn:bc3} (that we call BT), the representation will take one of the values $\{-2, -1, 0, 1, 2\}$. Hence, an ideal scale value must be between the lower bound (i.e., $(\alpha_1 - \alpha_2)$) and upper bound (i.e., $0.5\cdot(\alpha_1 + \alpha_2)$, $0.5$ is because of the maximum binary range which is $2$). Intuitively, the mean value of these scales yields the optimum scale value as given in Eq.~\ref{eqn:scale}. 

\vspace{-0.3cm}
\subsection{Transition Regularization}
\label{ssec:weightreg}
\vspace{-0.1cm}

As explained in~\cite{li2016ternary}, even if full-precision parameters are sampled from either a normal or uniform distribution, the allowable parameter range can still be large. A high approximation error is therefore expected. Furthermore, parameter updates with binary-precision are problematic as accurate error propagation is nearly impossible and greatly overfits the parameters.

We propose a novel regularization term that forces parameters to concentrate around binary transitions and penalizes those that are far from those points. For clarity, transition points correspond to $0$ for binary-weight models, $\{-\Delta, \Delta\}$ for ternary-weight models and $\alpha_{opt}$s for our models. Hence, the effects of parameter changes for binary-precision can be learned more quickly with this regularization term at train time. 

To favor these changes, a simple and effective trick is used. Mathematically, full-precision parameters $\mathbf{W}$ are corrupted with a random noise $\mathbf{\tilde{W}}=\mathbf{W} + 0.1 \cdot \mathcal{N}(0, \alpha)$. Later, these noisy weights are also quantized $\mathbf{\tilde{W}}^b$ similarly with Eq.~\ref{eqn:bc1}. Note that at the boundaries of transition points, binarized-representations for corrupted parameters differ from the original versions:   
\begin{eqnarray}
\label{eqn:bc5}
\mathcal{L} = \mathcal{L}_{pb} -\lambda ||\mathbf{W}^b - \mathbf{\tilde{W}}^b||_1
\end{eqnarray}

\begin{figure}
\centering
\vspace{-0.3cm}
\includegraphics[scale=1]{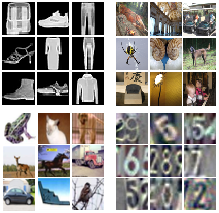}

\vspace{-0.3cm}
\caption{Visual samples from Fashion (Upper-Left), ImageNet2012 (Upper-Right), Cifar10 (Lower-Left) and SVHN (Lower-Right) datasets.}
\vspace{-0.3cm}
\label{fig:f1}
\end{figure}

\noindent Here, $\lambda$ is a coefficient that scales the contribution of regularization term to the overall loss function with a task-specific loss function $\mathcal{L}_{pb}$. Note that this coefficient is empirically set to $0.1$. From this equation, the absolute distance between actual and corrupted binary-precision weights is expected to be maximized. Hence, full-parameters are pushed close to these transition points. This notion is similar to Dropout~\cite{srivastava2014dropout} regularization where parameter space is intentionally corrupted.

In the experiments, we will show that this term is particularly useful for classification tasks than inverse problems. The main reason is that as explained in~\cite{erin2015deep}, regression tasks can lead to holes due to continuous solution spaces, so that sparsity cannot be maintained correctly in such problems.

\vspace{-0.3cm}
\subsection{Implementation Detail}
\label{ssec:weightreg}
\vspace{-0.1cm}

For the implementation detail\footnote{https://github.com/savasozkan/BofT}, each model is trained with Adam stochastic optimizer at train time. Mini-batch size is set 128, and learning rate scaling is adopted. In the classification task, VGG-6 (K) with batch normalization is used for acceleration and parameter regularization (i.e., 2×(K-C3) + MP2 +2×(2×K-C3) + MP2 + 2×(4×K-C3) + MP2 + 8×K-FC + Softmax and K is the number of filters). Here, "K-C3" is convolution layer with K filters of a $3\times3$ kernel. Also, "MP2" and "K-FC" denote a max pooling with stride $2$ and a fully-connected layer with K filters respectively. Similarly, espcn architecture with skip-connection is utilized for super-resolution and denoising tasks.

\vspace{-0.3cm}
\section{Experiments}
\label{sec:experiments}
\vspace{-0.3cm}

Experiments are conducted on two sets of tasks as visual classification and visual inverse problems to show the superiority of our method. The performance of our model (BT) is compared with binary-weight (BWN)~\cite{rastegari2016xnor}, ternary-weight (TWN)~\cite{li2016ternary}, trained ternary quantization (TTQ)~\cite{zhu2016trained} and LQ-Nets (LQ)~\cite{zhang2018lq} models that are selected as baselines. Note that all results are obtained by ours running on their publicly available codes. For classification tasks, the mean average precision (MAP)@1 metric is used, while PSNR scores are calculated for super-resolution and denoising tasks. For clarity, two problem sets are explained in two individual sections.

\vspace{-0.3cm}
\subsection{Visual Classification}
\label{ssec:expclass}

\begin{table}[t]
\begin{center}
\vspace{-0.1cm}

\caption{Comparisons with baselines on ImageNet2012 (INet), Fashion, Cifar10 and SVHN datasets. MAP@1 scores are reported. Best results for binary precision are bold.}
\scalebox{0.92}{
\begin{tabular}{c|c|c*{4}{c}|c}
\hline \hline
& Arch. & Full & BWN & TWN & TTQ & LQ & BT(ours) \\
\hline

\parbox[t]{1mm}{\multirow{2}{*}{\rotatebox[origin=c]{90}{INet}}} 

& K=64 & 42.81 & 35.30 & 38.25 & 37.76 & 39.01 & \textbf{40.21} \\
& K=128 & 44.13 & 40.08 & 42.59 & 41.18 &  42.65 & \textbf{42.80} \\

\hline

\parbox[t]{1mm}{\multirow{4}{*}{\rotatebox[origin=c]{90}{Fashion}}} 

& K=16 & 93.48 & 92.51 & 93.20 & 93.32 & 93.34 & \textbf{93.41} \\
& K=32 & 94.16 & 93.28 & 93.97 & 93.91 & \textbf{94.04} & 94.03 \\
& K=64 & 94.56 & 94.16 & 94.30 & 94.18 & 94.29 & \textbf{94.55} \\
& K=128 & 94.58 & 94.39 & 94.32 & 94.34 & 94.45 & \textbf{94.51} \\

\hline

\parbox[t]{1mm}{\multirow{4}{*}{\rotatebox[origin=c]{90}{Cifar10}}} 

& K=16 & 87.62 & 78.70  & 82.94 & 83.83 & \textbf{84.67} & 84.61 \\
& K=32 & 90.64 & 86.45 & 88.37 & 88.53 & 88.93 & \textbf{89.49} \\
& K=64 & 92.89 & 90.35 & 91.67 & 91.02 & 91.90 & \textbf{92.01} \\
& K=128 & 93.58 & 92.06 & 92.39 & 92.23 & 92.85 & \textbf{92.92} \\

\hline

\parbox[t]{1mm}{\multirow{4}{*}{\rotatebox[origin=c]{90}{SVHN}}} 

& K=16 & 94.14 & 93.15  & 93.44 & 93.32 & 94.04 & \textbf{94.09} \\
& K=32 & 94.93 & 94.32 & 94.84 & 94.69 & 94.83 & \textbf{94.88} \\
& K=64 & 95.43 & 95.10 & 95.22 & 95.29 & 95.31 & \textbf{95.37} \\
& K=128 & 95.76 & 95.17 & 95.28 & 95.65 & \textbf{95.84} & 95.74 \\

\hline \hline
\end{tabular}}
\label{tab:t2}
\vspace{-0.7cm}

\end{center}
\end{table}

Experimental results are reported on Fashion, ImageNet2012 (64×64), Cifar10 and SVHN datasets. Fashion, Cifar10 and SVHN datasets have 10 object/character classes, while ImageNet2012 has 1000 object classes. Visual samples for these datasets are illustrated in Fig.~\ref{fig:f1}. Random crop, padding and cutout policies are exploited as data augmentation. Moreover, softmax cross-entropy loss is used as the task-specific loss function $\mathcal{L}_{pb}$. Note that all trainable parameters, excluding fully-linear layers, are binarized in our paper.  

First, we compare our method with baseline models, and results are illustrated in Table~\ref{tab:t2}. As observed, our method outperforms all its binary-precision counterparts. This result validates that it improves the quantization performance by considering the approximation error. Especially for ImageNet2012 (INet), the increase is significant, and the proposed model yields close results with the full-precision architecture.  Moreover, the performance saturates when the number of parameters increases for all-binarized models as expected.

\begin{table}
\begin{center}

\caption{Impact of transition regularization term on ImageNet2012 (INet), Fashion, Cifar10 and SVHN datasets. MAP@1 scores are reported.}

\scalebox{0.92}{%
\begin{tabular}{c|c|c|c*{2}{c}}
\hline \hline
& Arch. & Full & BWN & TWN & BT(ours) \\
\hline

\parbox[t]{1mm}{\multirow{2}{*}{\rotatebox[origin=c]{90}{INet}}} 

& K=64 & 42.81 & 40.29 & 38.45  & 40.97 \\
& K=128 & 44.13 & 35.60 & 42.82 & 43.22 \\

\hline

\parbox[t]{1mm}{\multirow{4}{*}{\rotatebox[origin=c]{90}{Fashion}}} 

& K=16 & 93.48 & 93.11 & 93.36 & 93.45 \\
& K=32 & 94.16 & 93.76 & 94.08 & 94.12 \\
& K=16 & 94.56 & 94.34 & 94.42 & 94.55 \\
& K=32 & 94.58 & 94.45 & 94.50 & 94.55 \\

\hline

\parbox[t]{1mm}{\multirow{4}{*}{\rotatebox[origin=c]{90}{Cifar10}}} 

& K=16 & 87.62 & 79.42 & 83.42 & 85.03 \\
& K=32 & 90.64 & 86.78 & 88.52 & 90.12 \\
& K=16 & 92.89 & 90.57 & 91.92 & 92.12 \\
& K=32 & 93.58 & 92.26 & 92.57 & 92.99 \\

\hline

\parbox[t]{1mm}{\multirow{4}{*}{\rotatebox[origin=c]{90}{SVHN}}} 

& K=16 & 94.14 & 93.67 & 93.82 & 94.12 \\
& K=32 & 94.93 & 94.52 & 94.90 & 94.88 \\
& K=16 & 95.43 & 95.28 & 95.34 & 95.40 \\
& K=32 & 95.76 & 95.48 & 95.54 & 95.72 \\

\hline \hline
\end{tabular}}
\label{tab:t3}
\vspace{-0.7cm}

\end{center}
\end{table}

Finally, the impact of transition regularization term is reported in Table~\ref{tab:t3}. Results confirm that this term improves all threshold-based binarized weight models. In particular, these improvements reach up to $1\%$ for the shallow networks (i.e., K is lower) compared to results in Table~\ref{tab:t2}. The reason is that when the number of parameters increases, parameters are automatically concentrated in a narrow range, which has a positive impact on the quantization. 

\vspace{-0.3cm}
\subsection{Visual Inverse Problems}
\label{ssec:expres}
\vspace{-0.1cm}

Experiments are carried out on two well-known inverse problem tasks: visual super-resolution and visual denoise. At train time, images from the COCO val2017 dataset are used for both tasks. For high resolution, these images are rescaled by 2, 3, and 4 to simulate changes in resolution. At test time, the data sets Set5, Set14, Urban, and BSD100 are used. Additionally, different sigma values (i.e., with a mean of zero) are used to attach images for the denoise task. Tests are also performed on Set5 and Set14 records.

\begin{table}
\begin{center}

\caption{Super-resolution results on different datasets with various scale factors. PSNR scores are reported.}

\scalebox{0.92}{%
\begin{tabular}{c|c|c*{3}{c}|c}
\hline \hline
 & Dataset & Full & Bicubic & BWN & TWN & BT(ours) \\
\hline

\parbox[t]{1mm}{\multirow{4}{*}{\rotatebox[origin=c]{90}{Scale 2}}} 

& Set5 & 36.61 & 33.68 & 36.00 & 36.03 & 36.63  \\
& Set14 & 32.58 & 30.24 & 32.14 & 32.13 & 32.42  \\
& Urban & 29.51 & 26.61 & 28.91 & 28.83 & 29.31  \\
& BSD & 31.48 & 29.48 & 31.08 & 31.13 & 31.28  \\

\hline

\parbox[t]{1mm}{\multirow{3}{*}{\rotatebox[origin=c]{90}{Scale 3}}} 

& Set5 & 33.11 & 30.43 & 32.61 & 32.67 & 32.95  \\
& Set14 & 29.46 & 27.54 & 29.13 & 29.12 & 29.32  \\
& BSD &  28.38 & 27.14 & 28.21 & 28.25 & 28.34  \\

\hline

\parbox[t]{1mm}{\multirow{4}{*}{\rotatebox[origin=c]{90}{Scale 4}}} 

& Set5 & 30.82 & 28.45 & 30.35 & 30.41 & 30.72  \\
& Set14 & 27.70 & 26.01 & 27.38 & 27.43 & 27.66  \\
& Urban & 24.81 & 23.12 & 24.38 & 24.49 & 26.69  \\
& BSD &  27.01 & 25.91 & 26.73 & 26.81 & 26.97  \\

\hline \hline
\end{tabular}}
\label{tab:t4}

\vspace{-0.7cm}

\end{center}
\end{table}

First, an input image is converted to YUV color space for both tasks, and Y channel is used for both train and test phases. Also, our NN architecture is fully convolutional. It takes an input image and applies consecutive convolution layers with residual-based learning, as explained in~\cite{zhang2017beyond}. At the output, it is upsampled with a subpixel layer~\cite{shi2016real} for super-resolution (i.e., (64-C3)  + 2×(64-C3) + 2×(64-C3) + SubPixel + (1-C3)). Note that the first and last layers are full-precision networks, while the rest is binarized. 

Experimental results for super-resolution are reported in Table~\ref{tab:t4}. Our method outperforms all other baselines. In particular, for small scale factors, negligible performance drops can be observed compared to full-precision networks. Similarly, the reason is that the allowable parameter range is expected to be small for small scale factors, and the sparsity is decreased. 

\begin{table}
\begin{center}

\caption{Denoising results on different datasets with various sigma values. PSNR scores are reported.}

\scalebox{0.92}{%
\begin{tabular}{c|c|c*{3}{c}|c}
\hline \hline
 & Dataset & Noisy & Full & BWN & TWN & BT(ours) \\
\hline

\parbox[t]{1mm}{\multirow{2}{*}{\rotatebox[origin=c]{90}{0.05}}} 

& Set5 & 26.20 & 33.58 & 33.06 & 33.03 & 33.59  \\
& Set14 & 26.14 & 32.41 & 31.62 & 31.64 & 32.37  \\

\hline

\parbox[t]{1mm}{\multirow{2}{*}{\rotatebox[origin=c]{90}{0.1}}} 

& Set5 & 20.26 & 29.73 & 29.63 & 29.42 & 29.71  \\
& Set14 & 20.17 & 28.89 & 28.01 & 28.07 & 28.91  \\

\hline

\parbox[t]{1mm}{\multirow{2}{*}{\rotatebox[origin=c]{90}{0.2}}} 

& Set5 & 14.65 & 26.41 & 26.05 & 25.53 & 26.55  \\
& Set14 & 14.37 & 25.19 & 24.60 & 24.29 & 25.04  \\

\hline \hline
\end{tabular}}
\label{tab:t5}

\vspace{-0.7cm}

\end{center}
\end{table}

Denoising results are also shown in Table~\ref{tab:t5}. From the results, the proposed method achieves slightly better performance even for full-precision networks. Residual-based learning eventually promotes robustness with binarized networks since parameter range (i.e., additive noise varies $[0.05, 0.2]$) is heavily reduced. Hence, binarized weights can converge better solutions compared to other tasks. 

As indicated, the proposed regularization term yields similar or slightly worse performance for inverse problems. The main reason is that the learning process may lead to holes in continuous feature space partitioning.  

\vspace{-0.3cm}
\section{Conclusion}
\label{sec:conc}
\vspace{-0.3cm}

This paper proposes a novel model for binarized weight networks that achieves state-of-the-art performance compared to its counterparts. This model estimates a binary representation by taking into account the approximation error with an additional term. In addition, a novel regularization term is presented. This term is useful for threshold-based binarized models in particular. This term favors parameters that focus on binary transitions so that overfitting the parameters for binarized weights is avoided. In the experiments, we prove that this term is significant for classification-related tasks and enhances the performance of binarized methods. Experiments with different benchmarks for different tasks are conducted to confirm the superiority of our contributions.

\vspace{-0.3cm}

\bibliographystyle{IEEEbib}
\bibliography{Template}

\end{document}